\documentclass[runningheads]{llncs}
\usepackage{hyperref}
\usepackage{makecell}
\usepackage{graphicx}
\usepackage{amsmath}
\usepackage{amsfonts}
\usepackage{mathtools}
\usepackage{multirow}
\usepackage{comment}

\usepackage{color}
\newcommand{\ci}[1]{\textcolor{black}{#1}}
\newcommand{\mo}[1]{\textcolor{black}{#1}}

\DeclareMathOperator*{\argmin}{arg\,min}
\newcommand{\addpic}{9em}
\newcommand{\addpich}{5em}
\begin{document}
\title{Knowledge distillation for semi-supervised domain adaptation}

\authorrunning{H.~M.~Orbes-Arteaga et al.}
\author{Mauricio Orbes-Arteaga\inst{1,2,3,4} 
\and 
Jorge Cardoso\inst{4}  
\and 
Lauge S{\o}rensen\inst{1,2,3} 
\and 
Christian Igel\inst{1} 
\and 
Sebastien  Ourselin\inst{4} 
\and 
Marc Modat\inst{4}  
\and Mads Nielsen\inst{1,2,3} 
\and Akshay Pai\inst{1,2,3}}
\institute{%
Department of Computer Sceince, University of Copenhagen, Denmark
\and 
Cerebriu A/S, Copenhagen, Denmark
\and
Biomediq A/S, Copenhagen, Denmark
\and
King’s College London, United Kingdom
}

\maketitle              
\begin{abstract}
In the absence of sufficient data variation (e.g., scanner and protocol variability) in annotated data, deep neural networks (DNNs) tend to overfit during training. As a result, their performance is significantly lower on data from unseen sources compared to the performance on data from the same source as the training data. Semi-supervised domain adaptation methods can alleviate this problem by tuning networks to new target domains without the need for annotated data from these domains. Adversarial domain adaptation (ADA) methods are a popular choice that aim to train networks in such a way that the features generated are domain agnostic. However, these methods require careful dataset-specific selection of hyperparameters such as the complexity of the discriminator in order to achieve a reasonable performance. We propose to use knowledge distillation (KD) -- an efficient way of transferring knowledge between different DNNs -- for semi-supervised domain adaption of DNNs. It does not require dataset-specific hyperparameter tuning, making it generally applicable. The proposed method is compared to ADA for segmentation of white matter hyperintensities (WMH) in magnetic resonance imaging (MRI) scans generated by scanners that are not a part of the training set. Compared with both the baseline DNN (trained on source domain only and without any adaption to target domain) and with using ADA for semi-supervised domain adaptation, the proposed method achieves significantly higher WMH dice scores. 

\end{abstract} 

\section{Introduction}
In the presence of a large training dataset that covers all possible data variations, deep neural networks (DNNs) can achieve super-human performance in 
image recognition and semantic segmentation tasks. However, in medical image segmentation tasks large annotated training datasets are often scarce. In addition, training and test data are drawn from different distributions. For example, the images were obtained  using different scanners at different sites 
or the demographics of the subjects differ. This violation of the i.i.d.\ assumption (i.e., that training and test data are drawn independently from the same distribution) typically has the effect that the performance on the test data is significantly worse than on the training data.


Domain adaptation (DA) approaches try to alleviate the problem of applying models in new domains with different characteristics. In particular, semi-supervised DA methods provide a way to learn structure from unlabeled data in new domains. Among the several semi-supervised DA (SSL-DA) methods proposed, the most popular one is  \textit{adversarial training based domain adaptation} (ADA). ADA relies on generating features that are invariant with respect to a domain discriminator. ADA requires extensive parameter optimization due to the necessity of a robust discriminator. \mo{And a recent study pointed out the flaws in the evaluation of SSL-DA methods} \cite{oliver2018realistic}. 

In this paper, we evaluate a modified \emph{knowledge distillation} (KD)~\cite{hinton2015distilling,lopez2015unifying} method for generalizing DNNs to new domains with a common clinical problem in contrast to using ADA methods. The datasets chosen for evaluation not only involve different magnetic resonance images (MRIs), but also were acquired on subjects with different demographic makeup. Through our evaluation, we show that the proposed KD is generally able to achieve better dice scores in segmenting white matter hyperintensities (WMH) on datasets that are not a part of the training data and do not share any attributes when compared to baseline and ADA. 

\section{Related work}
Among the recent works on DA, several methods rely on using a small amount of data (\emph{annotated}) to fine-tune a baseline model \cite{hoffman2013efficient,karani2018lifelong}. The performance of this approach not only relies on a new -- albeit small -- set of annotations but also on the choice of the set. In contrast, SSL-DA do not use data annotations on new target domains. Adversarial training is a popular  SSL-DA method \cite{tzeng2017adversarial,sun2016deep,hoffman2017cycada}. Here, networks are trained in such a way that the generated features are agnostic to the data domain with respect to a domain discriminator. A similar solution, ADA, was employed by \cite{kamnitsas2017unsupervised} to adapt networks to be agnostic to domain changes.

Another class of DA method use KD to transfer representations between data domains. For instance,~\cite{gupta2016cross} proposed using KD to transfer knowledge between different modalities of the same scene. Closely related to our work is \cite{huang2018omni}, where the authors propose to use omni-supervised learning (OSL) to include unlabelled data in the learning process. Here, data distillation is used to generate an ensemble of predictions from multiple transformations of unlabeled data, using a teacher model, to generate new training annotations. The proposed method differs from this method on two accounts: a) Only soft labels are used to train the single student network, \mo{where the idea is to improve segmentation by learning label similarities from unannotated data} b) the data included in the training of the student involves data from new domains in small amounts in contrast to OSL. 

\section{Methods}
In SSL-DA methods, we assume the source domain images and their annotations, $(x_s, y_s) \in \mathbf{X}_s$, are drawn from a  distribution $p_s(x_s,y)$. The target domain images $x_t \in \mathbf{X}_t$, are drawn from a distribution $p_t(x_t,y)$ where there are no annotations available. 
\ci{We consider classification into $K$ classes.}
In an ideal scenario, where $p_s$ and $p_t$ are sufficiently similar, the goal is to find a feature representation mapping $f$ \ci{that maps an input to $K$ scores, where the $i^{\text{th}}$
score models (up to a constant) the logarithm of the probability that the input belongs to class $K$.}
\ci{These scores can then} be mapped by $\sigma: \mathbb{R}^{K} \rightarrow \mathbb{R}^{K}$ to probability maps 
\ci{over the} classes. 
SSL-DA 
first finds a function $f_s$ \ci{performing well} on 
\ci{a source}
domain and then 
\ci{finds} 
a new $f_t$ based on $f_s$ that 
\ci{performs well}  on the target domain. Vanilla supervised learning methods rely on including annotations from both $\mathbf{X}_s$ and $\mathbf{X}_t$.

In the popular ADA method, the goal is to minimize the distance between the empirical distributions of $p_s(f_s(\mathbf{X}_s)|y)$ and $p_t(f_t(\mathbf{X}_t)|y)$. \mo{Here,  a discriminator $D$  is a neural network that distinguishes between the two domains. Therefore,  the discriminator acts as a discrepancy measure that brings the two distributions together. Overall, adversarial training involves train a network that generates $f$ in a standard supervised manner that  is indistinguishable by a discriminator}~\cite{tzeng2017adversarial,kamnitsas2017unsupervised}.

\subsection{Knowledge distillation for Domain adaptation}
\label{KD}
KD~\cite{hinton2015distilling} was originally intended to compress neural networks with high number of parameters with networks of lower complexity. The objective is to teach a simpler student network to imitate a more complex trained teacher network, through a loss function called the distillation loss. 
To perform unsupervised domain adaptation, we proposed to use the teacher/student learning strategy. Specifically, the data from the source domain is used to train a teacher model in a supervised fashion. Then, the trained teacher is used to generate posterior probability maps or soft labels on the union of source and target data. These posterior probabilities are used instead of usual hard labels to train the student or target model. Note, this approach can take advantage of large amounts of unlabeled  data acquired from any number of domains. An attractive feature of distillation loss is the soft representation of one-hot encoded label vectors which allow the student to be optimized over a smoother optimization landscape. Moreover, the smooth representation of labels  also allows the learning of label similarities, which is particularly useful in learning boundaries in semantic segmentation tasks. The proposed semi-supervised learning method is formulated below.

\textbf{Training the teacher or source domain model:}
Consider a set of  $N$  manually annotate images from  a source domain $\mathbf{X}_s = \{ (x_i,y_i), i=1 \dots N \}$, where $x_i \in \mathbb{R}^d$ represent a $d$-dimensional MR scan, with $v=1 \dots V$ voxels, and $y_i \in [0,1]^K$ with $\|y_i\|_1=1$ its correspondent label. Assuming  there is a set $F_s$ that holds functions $f:\mathbb{R}^d \rightarrow \mathbb{R}^K$  we aim to learn a feature representation $f_s$ (teacher model) which follows the  optimization of a loss function, $l$, according to Equation~~\eqref{standard}
\begin{align}
    \argmin_{f\in F_s} &\frac{1}{N} \sum_{x_i\in \mathbf{X_s}}l( y_i, \sigma(f_s(x_i))) \label{standard}\\
    [\sigma(z)]_{k} &= \frac{\mathrm{e}^{[z]_{k}}}{\sum_{l=1}^{K} \mathrm{e}^{[z]_l}} 
\end{align}
In a standard supervised learning way, the teacher network is optimized using the cross-entropy loss function (or any differentiable loss function of choice).

\textbf{Training the student or target model:} Even though $f_s$ is suitable to segment 
\ci{the images from the source domain}
$\mathbf{X}_s$, it may not be suitable for data coming from a different data distribution $\mathbf{X}_t$. Our goal is find a function $f_t \in F_t$, which is suitable to segment data from $\mathbf{X}_t$.  Assuming, we have access to a limited set of unlabeled scans in the target domain $\mathbf{X}_t=\{x_i, i= 1 \dots M\}$, we can then create a set 
\begin{align*}
\mathbf{X}_U =  \{(x_i, y_i)  \,|\, x_i \in \mathbf{X}_s, y_i=f_s(x_i), 1\le i \le N \}  \cup \\
\{(x_i, y_i)  \,|\, x_i \in \mathbf{X}_t, y_i=f_s(x_i), 1\le i \le M \}
\end{align*}
that may be used to optimize a student using the distillation loss. 
\mo{Through soft-representations of this union dataset, the student is expected to learn a better mapping to the labels than the teacher network. When training the student network, we consider probability distributions over the labels as targets, not single classes. This representation reflects the uncertainty of the prediction by the teacher network}. \ci{The function $f_t$ is found by (approximately) solving},
\begin{equation}
    \argmin_{f\in F_t} \frac{1}{(N+M)} \sum_{x_i\in\mathbf{X_U}}l( \sigma(T^{-1}f_s(x_i)), \sigma(f_{t}(x_i)))\enspace,
\end{equation}
Here, $T > 1$ is the temperature parameter which controls the softness of the class probability prediction given by $f_s$.

\section{Experiments and Results}
\subsection{Databases}
The \textbf{WMH segmentation challenge}
\begin{table}[ht]
    \caption{Summary of data characteristics in the WMH challenge database
    \label{demo}}
    \centering
    \begin{tabular}{lc@{\quad}c@{\quad}c@{\quad}cc}
    \hline\noalign{\smallskip}
     Clinic & Scanner Name & Voxel Size($m^3$) & Size & $\#$ of images \\
     \noalign{\smallskip}
\hline
\noalign{\smallskip}
     Utrech   &  3T Philips Achieva &  $ 0.96 \times 0.95 \times 3.00 $ & $240 \times 240 \times 48$ & 20 \\
     Singapore &  3T Siemens TrioTim &  $ 1.00 \times 1.00 \times 3.00 $ & $252 \times 232 \times 48$ & 20 \\
     Amsterdam      &  3T GE Signa HDxt   &  $ 1.20 \times 0.98 \times 3.00 $ & $132 \times 256 \times 83$ & 20 \\ 
     \noalign{\smallskip}
\hline
     \end{tabular}
\end{table}
(\hyperlink{https://wmh.isi.uu.nl/}{https://wmh.isi.uu.nl/} ) dataset is a public database that contains T1-weighted and FLAIR scans for 60 subjects from three different clinics. The data also consists of manual annotations of WMH from presumed vascular origin. T1-weighted images have been registered to FLAIR since annotations were performed in this space. The images were also corrected for bias field inhomogenities using SPM12. An important feature of this dataset is that the scanners and demographics have variance as show in the Table~\ref{demo}. 
\subsection{Experimental setup}
\label{cross}
One of the main objectives of the paper is to use semi-supervised learning to perform domain adaptation. We use the WMH challenge dataset to perform cross-clinical experiments in segmenting WMH on FLAIR images. We consider several scenarios to establish the performances of ADA and KD. The scenarios are described below. Note that, to evaluate the performance of the algorithms, dice overlap measures are used throughout.
\begin{itemize}
    \item Lower bound baseline, \textbf{L-bound}: Here a baseline DNN model is trained on the source dataset to establish a lower bound performance. The DNN is trained on the source domain images henceforth referred to as \textbf{S}, and tested on 20 subjects from a target dataset~\textbf{T}. 
    \item Upper bound baseline, \textbf{U-bound}: Here, a baseline DNN model is trained like L-Bound, however, the training dataset is a union of images from both \textbf{S} and a subset of \textbf{T} (10 subjects, with annotations). The network is evaluated on the remaining 10 subjects in \textbf{T}.
    \item Adversarial domain adaptation, \textbf{ADA}: Following~\cite{kamnitsas2017unsupervised}, we attempt at training a DNN model that is invariant to data domains. In this paper, to be consistent with KD, we train the domain discriminator based on the final layer of the baseline, in contrast to what was proposed in~\cite{kamnitsas2017unsupervised}. We use a discriminator composed of 4 convolutional layers with 8, 16 32, 64 number of filters, followed by 3 fully connected layers with 64, 128 and 2 neurons. For this experiment, like U-bound, the training dataset is a union of images from both \textbf{S} and a subset of \textbf{T} (10 subjects, without annotations). The network is evaluated on the remaining 10 subjects in \textbf{T}.
    \item Knowledge distillation, \textbf{KD}: The experimental setup for KD is the same as ADA. A temperature of 2 is used in the softmax for the distillation loss. The student network trained is identical to the teacher network whose architecture is a standard UNet (like L-bound, U-bound, and ADA) optimized with an ADAM loss function and a learning rate of $10^{-4}$ with is gradual decrease after epoch 150. The network is trained for 400 epochs. 
    \item{Adaptation on-the-fly}: A clinically relevant scenario is adapting to a small set of test images on the fly by keeping the teacher/baseline model constant. To validate this scenario, we apply ADA and KD on the same 10 unannotated \textbf{T} that are included in the training, but subject-wise. In other words, separate adaptation is performed on each instance of \textbf{T}, instead of including them together.
\end{itemize}

\subsection{Results}
Various combinations of mismatched (in terms of clinics) training and testing data were used. For instance, if the training data is from clinic 1 (Utrecth), the testing data is from either clinic 2 (Singapore), or clinic3 (Amsterdam). 
We did not test on two different clinics even though this scenario is practical. 
Table~\ref{table-cross-clinic} illustrates mean dice coefficients (two folds) for each of the scenarios mentioned in Section~\ref{cross} except for adaptation on the fly which is illustrated in Table~\ref{fly}. KD outperformed ADA in nearly all scenarios except for domain adaptation from Singapore clinic to Utrecht clinic and vice versa. For domain adaptation from Utrecht clinic to Singapore clinic, ADA was significantly better than KD. 
In the vice-versa situation, KD 
achieved a better mean
which is statistically not significant 
In all other scenarios, KD 
yielded statistically better dice overlaps compared to ADA. Note that the statistical comparison are made only between ADA and KD. 
\begin{table}[ht]
    \caption{ Illustrates dice overlaps (with variance). Bold fond indicates statistical significance at $5\%$, p-values   (paired-sample t-test at was used to computed p-values, which were $0.0002 < p < 0.02$). Only ADA and KD methods are considered in the statistical comparison.
    \label{table-cross-clinic}}
    \centering
    \begin{tabular}{l | c | c | c | c |} 
          \diaghead{\theadfont Diag Columnmn}%
  {Training}{Test} & Method & \ Utrech  \ & \ Singapore \ & \ \ Amsterdam \ \ \\
         \hline
         \multirow{ 4 }{*}{Utrech}& L-bound     &            & 0.6126 ( 0.1092)           & 0.7207  (0.0793) \\
         & ADA                                   &            & \textbf{0.7004 ( 0.1057)}  & 0.7144  (0.0968) \\
         & KD                                    &            & 0.6456 ( 0.0905)     & \textbf{0.7548 (0.0755)} \\
         & U-bound                               &            & 0.8031 ( 0.1148)           & 0.7704 (0.0787)  \\
         \hline  
         \multirow{ 4 }{*}{Singapore}& L-bound   &  0.6693 ( 0.2271)        &                 &  0.7368 (0.0931)  \\
         & ADA                                   &  0.6859 ( 0.2036)        &                 &  0.7337 (0.0912) \\
         & KD                                    &  0.6924 ( 0.2103)        &                 &  \textbf{0.7499 (0.0877)} \\
         & U-bound                               &  0.7063 ( 0.2016)        &                 &  0.7699 (0.0851)  \\
         \hline
         \multirow{ 4 }{*}{Amsterdam}& L-bound   &  0.6471 (0.2086)           & 0.6811 (0.1172)          &  \\
         & ADA                                   &  0.6800 (0.2128)           & 0.7202 (0.1154)          &  \\
         & KD                                    &  \textbf{0.6909 (0.2135)}  &\textbf{0.7482 (0.0975)}  &  \\
         & U-bound                               &  0.7208 (0.1851)           & 0.7988 (0.0869)          &  \\
         \hline
     \end{tabular}
\end{table}
In the adaptation-on-the-fly scenario, KD yields significantly better dice overlaps on a majority of the scenarios, the superior performance of ADA remains in the experiment that involves domain adaptation from Utrecht clinic to Singapore clinic. However, in the vice-versa scenario, KD performance better than ADA. To illustrate the differences in segmentations between KD and ADA, we plot the segmentations (scenario, Utrecht clinic to Amsterdam clinic) in Figure~\ref{qual}. As illustrated, both the methods perform quite well in segmenting lesions with relatively larger volume, however, the main difference is evident in segmenting smaller lesions, specially in the deep white matter regions. 
\begin{table}[ht]
     \caption{ Mean dice overlaps from the adaptation-on-the-fly scenario. Bold fond indicates statistical significance at $5\%$, p-values (paired-sample t-test at was used to computed p-values, which were $0.0003 < p < 0.04$). Only ADA and KD methods are considered in the statistical comparison.
    \label{fly}}
    \centering
    \begin{tabular}{l | c | c | c | c |}
          \diaghead{\theadfont Diag Columnmn}%
  {Training}{Test} & Method & \ Utrech  \ & \ Singapore \ & \ \ Amsterdam \ \ \\
         \hline
         \multirow{ 2 }{*}{Utrech}& KD     &            &     \textbf{0.6285 ( 0.097}   & \textbf{0.7465(0.0855)} \\
         & ADA                              &            &     0.7075 ( 0.095)        & 0.7220(0.0995) \\
         
         \hline  
         \multirow{ 2 }{*}{Singapore}& KD        & \textbf{0.6945(0.1825)}  &                 & 0.7425(0.0805)  \\
                                     & ADA                                   & 0.6680(0.1945)         &                 & 0.7370(0.0880)  \\
         \hline 
         \multirow{ 2 }{*}{Amsterdam}&  KD            & \textbf{0.6745 ( 0.2005)}  &  \textbf{0.7395 (0.1165)} &  \\
         & ADA                                   & 0.6625 ( 0.1890)           &  0.7100 (0.1125)  &  \\
         \hline
     \end{tabular}
\end{table}
It is interesting to note that the adaptation-on-the-fly and the classical scenarios yield nearly the same dice indicating a good generalisability and less dependency on the choice of the small dataset coming from the target domain.

\section{Discussion}
The main objective of this paper was to present domain adaptation from a semi-supervised learning perspective. We have evaluated a modified knowledge distillation approach and compared it to the popular adversarial approach under different clinical scenarios. Overall, the knowledge distillation approach gave better results and is relatively simpler to design when compared to the more architecture-dependent adversarial approaches. Adversarial approaches require extensive tuning of DNN architectures, especially for
the discriminator, in order to achieve reasonable performances. 
In contrast, KD  only involves choosing  the temperature parameter which can be chosen only based on the performances on the source domain. 
\begin{table}[ht]
\begin{tabular}{  c  c  c  c }
     Target & ADA & KD & U-bound\\
     \hline
     \includegraphics[width=\addpic,height=\addpich]{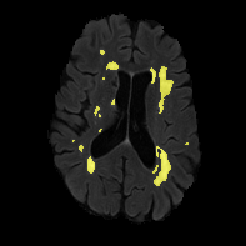} & \includegraphics[width=\addpic,height=\addpich]{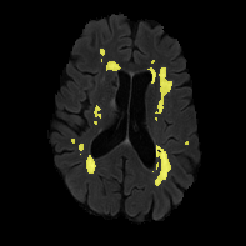} & \includegraphics[width=\addpic,height=\addpich]{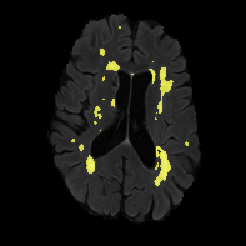} & \includegraphics[width=\addpic,height=\addpich]{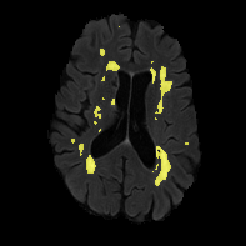} \\
           \includegraphics[width=\addpic,height=\addpich]{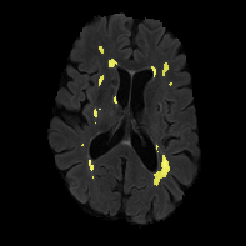} & \includegraphics[width=\addpic,height=\addpich]{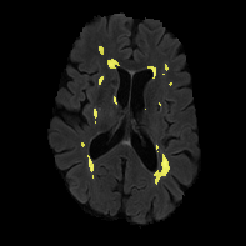} & \includegraphics[width=\addpic,height=\addpich]{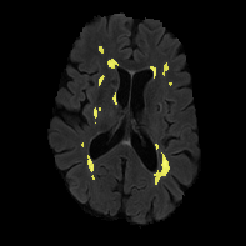} & \includegraphics[width=\addpic,height=\addpich]{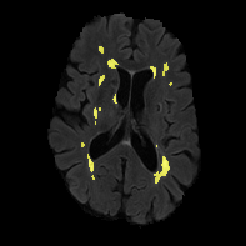} \\
     
\end{tabular}
\caption{Illustration of the segmentation's obtained with different methods trained on the Utrecht dataset and tested on the Amsterdam dataset. The top and bottom row illustrate segmentations on two different subjects.}
\label{qual}
\end{table}
One of the interesting outcomes is the inferior performance of KD on domain adaptation in scenario, Utrecht clinic to Singapore clinic. One of the reasons may be attributed to not just scanner differences but also differences in demographics. This may have led to an inferior teacher performance that the student network relies on. To verify this, we used the improved network from domain adaptation using ADA as a teacher and then trained a student based on it. We observed that the mean dice overlap improved from $0.65 \rightarrow 0.69$. 

In future work, we will consider combining the adversarial approaches with knowledge distillation to improve the generalisability of DNNs across domains without the need for large annotated datasets. 

\subsubsection{Acknowledgements}
This project has received funding from the EU H2020 under the Marie Sk\l odowska-Curie grant agreement No 721820. We would like to thank Microsoft Azure and NVIDIA for providing the necessary computational resources for the project.



\bibliographystyle{splncs}
\bibliography{mauricio}

\begin{thebibliography}{10}

\bibitem{oliver2018realistic}
Oliver, A., Odena, A., Raffel, C.A., Cubuk, E.D., Goodfellow, I.:
\newblock Realistic evaluation of deep semi-supervised learning algorithms.
\newblock In: Advances in Neural Information Processing Systems. (2018)
  3239--3250

\bibitem{hinton2015distilling}
Hinton, G., Vinyals, O., Dean, J.:
\newblock Distilling the knowledge in a neural network.
\newblock arXiv preprint arXiv:1503.02531 (2015)

\bibitem{lopez2015unifying}
Lopez-Paz, D., Bottou, L., Sch{\"o}lkopf, B., Vapnik, V.:
\newblock Unifying distillation and privileged information.
\newblock arXiv preprint arXiv:1511.03643 (2015)

\bibitem{hoffman2013efficient}
Hoffman, J., Rodner, E., Donahue, J., Darrell, T., Saenko, K.:
\newblock Efficient learning of domain-invariant image representations.
\newblock arXiv preprint arXiv:1301.3224 (2013)

\bibitem{karani2018lifelong}
Karani, N., Chaitanya, K., Baumgartner, C., Konukoglu, E.:
\newblock A lifelong learning approach to brain mr segmentation across scanners
  and protocols.
\newblock In: International Conference on Medical Image Computing and
  Computer-Assisted Intervention, Springer (2018)  476--484

\bibitem{tzeng2017adversarial}
Tzeng, E., Hoffman, J., Saenko, K., Darrell, T.:
\newblock Adversarial discriminative domain adaptation.
\newblock In: Proceedings of the IEEE Conference on Computer Vision and Pattern
  Recognition. (2017)  7167--7176

\bibitem{sun2016deep}
Sun, B., Saenko, K.:
\newblock Deep coral: Correlation alignment for deep domain adaptation.
\newblock In: European Conference on Computer Vision, Springer (2016)  443--450

\bibitem{hoffman2017cycada}
Hoffman, J., Tzeng, E., Park, T., Zhu, J.Y., Isola, P., Saenko, K., Efros,
  A.A., Darrell, T.:
\newblock Cycada: Cycle-consistent adversarial domain adaptation.
\newblock arXiv preprint arXiv:1711.03213 (2017)

\bibitem{kamnitsas2017unsupervised}
Kamnitsas, K., Baumgartner, C., Ledig, C., Newcombe, V., Simpson, J., Kane, A.,
  Menon, D., Nori, A., Criminisi, A., Rueckert, D.,  et~al.:
\newblock Unsupervised domain adaptation in brain lesion segmentation with
  adversarial networks.
\newblock In: International conference on information processing in medical
  imaging, Springer (2017)  597--609

\bibitem{gupta2016cross}
Gupta, S., Hoffman, J., Malik, J.:
\newblock Cross modal distillation for supervision transfer.
\newblock In: Proceedings of the IEEE conference on computer vision and pattern
  recognition. (2016)  2827--2836

\bibitem{huang2018omni}
Huang, R., Noble, J.A., Namburete, A.I.:
\newblock Omni-supervised learning: Scaling up to large unlabelled medical
  datasets.
\newblock In: International Conference on Medical Image Computing and
  Computer-Assisted Intervention, Springer (2018)  572--580

\end{thebibliography}

%
%
%
%





\end{document}